 \documentclass[tablecaption=bottom,wcp]{jmlr} 

\usepackage{booktabs}
\usepackage[load-configurations=version-1]{siunitx} 

 \usepackage{hyperref}
 \usepackage{caption}

\theorembodyfont{\upshape}
\theoremheaderfont{\scshape}
\theorempostheader{:}
\theoremsep{\newline}

\newcommand{\Bern}{\text{Bern}}
\newcommand{\KLD}[2]{
    D_{\mathrm{KL}}\left[ #1 \middle\| #2 \right]
}

\newcommand{\defeq}{\overset{\underset{\mathrm{def}}{}}{=}}

\newcommand{\Elbo}{\mathcal{L}}
\newcommand{\Normal}{\mathcal{N}}

\newcommand{\Prob}{p}
\newcommand{\Expect}{\mathbb{E}}

\newcommand{\reals}{\mbox{\(\mathbb R\)}}

\jmlrproceedings{AABI 2019}{2nd Symposium on Advances in Approximate Bayesian Inference, 2019}

\title[
    GP-ALPS: Automatic Latent Process Selection for MOGPs
]{
    GP-ALPS: Automatic Latent Process Selection for Multi-Output Gaussian Process Models 
}

 \author[Berkovich, Perim and Bruinsma]{\Name{Pavel Berkovich}\thanks{Invenia Labs, Cambridge, UK}\thanks{University College London} \Email{p.berkovich@cs.ucl.ac.uk}\\
 \Name{Eric Perim}\footnotemark[1] \Email{eric.perim@invenialabs.co.uk}\\
 \Name{Wessel Bruinsma}\footnotemark[1]\thanks{University of Cambridge} \Email{wpb23@cam.ac.uk}
}

\begin{document}
\maketitle

\vspace{-8mm}



\section{Introduction}
\label{sec:intro}
%

A principled approach to prediction tasks is to choose a statistical model that explains the data.
The choice of the \emph{model class} 
is crucial and has to observe the \textit{bias--variance trade-off}, which motivates the need for principled approaches to selecting the best model class from a set of options. Whilst model selection can be done manually by trial and error, the process tends to consume considerable time and resources and be prone to human biases.
Bayesian model selection \citep{mackay1992bayesian,kuo1998variable,rasmussen2001occam}, treats the model class as a \emph{random variable} and computes its posterior distribution. It offers a built-in complexity regulariser, commonly known as Bayesian Occam’s razor, which penalises models whose complexity is excessive or too modest. As a result, Bayesian model selection assigns high posterior probability to model classes whose complexity is ``just right’’.

Gaussian processes (GPs) are a popular and widely used approach to single-output nonlinear regression \citep{williams2006gaussian}.
They constitute a probabilistic modelling framework that is tractable, modular, and interpretable.
GPs can be extended to multiple output and have in this setting successfully been applied to problems as diverse as analysis of neuron activation patterns \citep{Yu:2009:Gaussian-Process_Factor_Analysis_for_Low-Dimensional}, image upscaling \citep{akhtar2016hierarchical}, and solar panels' output prediction \citep{Dahl:2018:Grouped_Gaussian_Processes_for_Solar}.
One of the simplest and most widely adopted approach to extend GPs to multiple outputs is to model each output as a linear combination of a collection of shared, unobserved latent Gaussian processes \citep{wackernagel1997multivariate}, henceforth referred to as the Linear Mixing Model (LMM).
A pressing issue with this approach is choosing the complexity of the latent space, which constitutes choosing the number of latent processes and their kernels.
These choices are typically done manually \citep{Teh:2005:Semiparametric_Latent_Factor,Osborne:2008:Towards_Real-Time_Information_Processing_of,Yu:2009:Gaussian-Process_Factor_Analysis_for_Low-Dimensional}, which can be time consuming and prone to overfitting.


In this work, we apply Bayesian model selection to the calibration of the complexity of the latent space.
We propose an extension of the LMM that automatically chooses the latent processes by turning off those that do not meaningfully contribute to explaining the data.
We call the technique Gaussian Process Automatic Latent Process Selection (GP-ALPS). The extra functionality of GP-ALPS comes at the cost of exact inference, so we devise a variational inference (VI) scheme and demonstrate its suitability in a set of preliminary experiments. We also assess the quality of the variational posterior by comparing our approximate results with those obtained via a Markov Chain Monte Carlo (MCMC) approach.

\section{Automatic Latent Process Selection (ALPS) for MOGPs}
\label{sec:alps}
We adopt the following formulation of the Linear Mixing Model: 
\begin{equation}
    x_j \sim \mathcal{GP}(0, k_j(t, t')), \quad
    f(t) = H x(t), \quad
    y_i(t) \sim \mathcal{N}(f_i(t), \sigma_i^2).
\end{equation}
In the LMM, $f_i(t) = \sum_{j=1}^m H_{ij} x_j(t)$ is a linear combination of unobserved processes $(x_j)_{j=1}^m$, where
we call $H$ the \emph{mixing matrix} and $x$ the \emph{latent processes}.
Our approach, named Gaussian Process Automatic Latent Process Selection (GP-ALPS), aims to automatically select those latent processes $x_j$ that meaningfully contribute to the observed signal.
It does so by multiplying every $x_j$ by a Bernoulli random variable $b_j$, which gives the model the ability to exclude $x_j$ from contributing to $f$: $f(t) = H(x(t)\circ b)$, where $\circ$ denotes the Hadamard product. This approach can be interpreted as a form of drop-out regularisation \citep{nalisnick2019dropout} on the latent processes. GP-ALPS also includes a prior over $H$.
In summary, GP-ALPS is given by the following generative model:
\begin{equation}
\begin{gathered}
    x_j \sim \mathcal{GP}(0, k_j(t, t')), \quad
    b_j \sim \operatorname{Bern}(\theta_j), \quad
    H_{ij} \sim \mathcal{N}(0, s_{ij}), \\
    f(t) = H (x(t) \circ b), \quad
    y_i(t) \sim \mathcal{N}(f_i(t), \sigma_i^2).
\end{gathered}
\end{equation}
Each of the $2^m$ possibilities for the vector $b$ identifies a model class, so the prior effectively describes an ensemble of $2^m$ different models, corresponding to all possible combinations of the latent functions. Another interpretation of GP-ALPS is that the latent processes are various features on which the observed signal can depend, which makes GP-ALPS a method to perform automatic feature selection. 



\section{Variational Inference Scheme}
\label{subsec:VI}

Let $Y \in \reals^{p \times n}$ denote observed data at input locations $t \in \reals^n$.
Augment the model with \emph{inducing variables} $X^z \in \reals^{m \times \ell}$ at input locations $t^z \in \reals^{\ell}$, which are assumed to be sufficient statistics for the latent processes \citep{titsias2009variational,Hensman:2013:Gaussian_Processes_for_Big_Data,Nguyen:2014:Collaborative_Multi-Output}.
To perform inference, we introduce a structured mean-field approximate posterior distribution
\[
    q(X, X^z, H, b)
    = p(X | X^z) q(X^z) q(H) q(b)
\]
where we choose $q(X^z)$, $q(H)$, and $q(b)$ by minimising the Kullback--Leibler divergence with respect to the true posterior, using stochastic gradient-based optimisation:
\[
    (q^*(X^z), q^*(H), q^*(b))
    = \operatorname{argmin}_{(q(X^z),q(H),q(b))}\KLD{q(X, X^z, H, b)}{p(X, X^z, H, b | Y)}
\]
We let $q(X^z)$ be a Gaussian that factorises over the latent processes and $q(H)$ a fully factorised Gaussian.
The approximate posterior $q(b)$, however, is troublesome, because $b$ is discrete, which means that we cannot just use the reparametrisation trick \citep{Kingma:2013:Auto-Encoding_VB, titsias2015local, rezende2014stochastic}.
We therefore let $q(b)$ be a continuous relaxation of the Bernoulli distribution called the \emph{concrete distribution} \citep{maddison2016concrete}.
To compute the ELBO, we also approximate $p(b)$ with the concrete distribution, as the cross-entropy $\Expect_{q(b)}\log p(b)$ would not be well-defined in the case in which $q(b)$ is continuous and $p(b)$ is discrete \citep{maddison2016concrete}.
For the temperature of the concrete distributions, we use a particular annealing scheme.
See Appendix \ref{apd:vi} for a more detailed description of the variational inference scheme.

To assess the quality of the variational approximate posterior, we compare it against Gibbs sampling, 
which has theoretical guarantees to converge to the true posterior in the infinite time limit.
The key insight is that $f$ is bilinear in $H$, $x$, which means that
$p(X | Y, H, b)$ and
$p(H | X, Y, b)$ are tractable (just Bayesian linear regression);
and $p(b_j | X, Y, H, b_{\neg j})$ is tractable because $b$ is discrete.
See Appendix \ref{apd:mcmc} for a more detailed description of the Gibbs sampler.

\section{Experimental Results}
\label{sec:experimental}


\subsection{Square Wave Decomposition} 
\label{subsec:square}

\begin{figure}[t]
   \subfigure[Wave reconstruction]{
       \label{fig1}
        \includegraphics[
            width=0.475\textwidth,
        ]{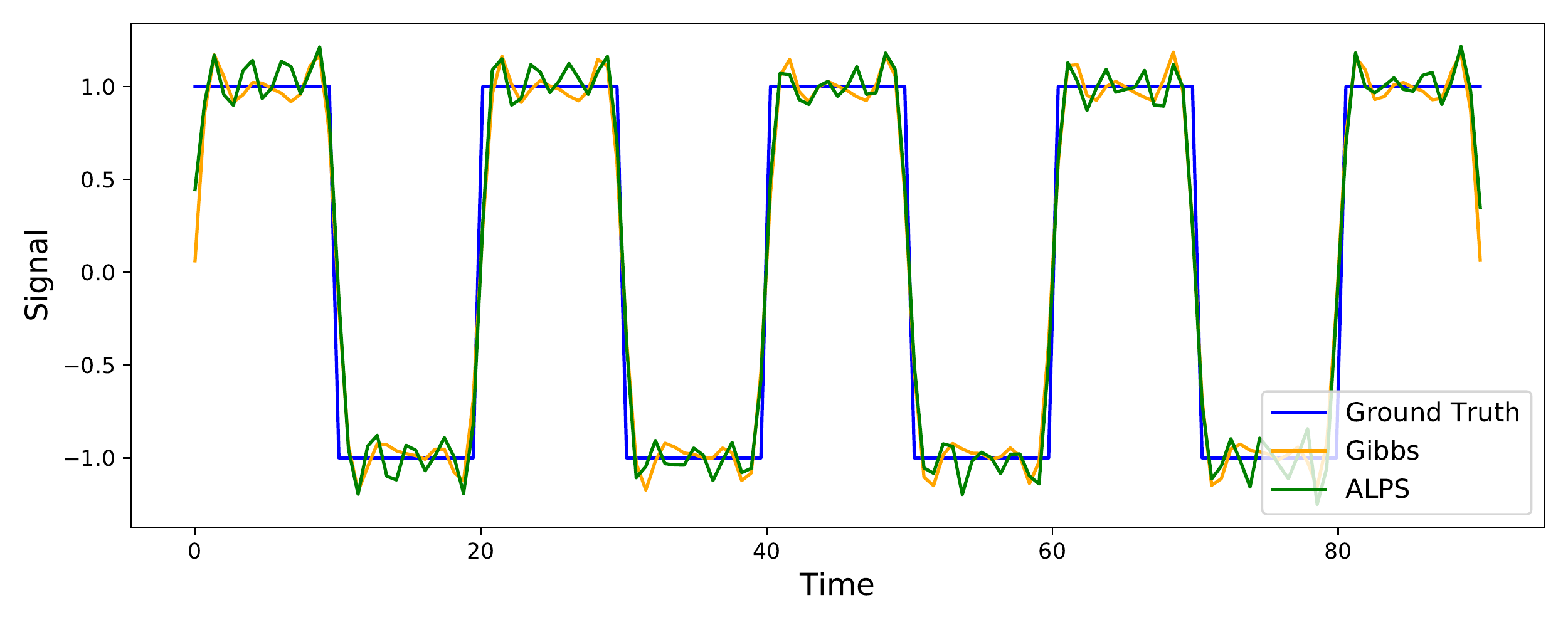}
   }
   \hfill
   \subfigure[Bernoulli activation probabilities]{
        \label{fig2}
        \includegraphics[
            width=0.475\textwidth,
        ]{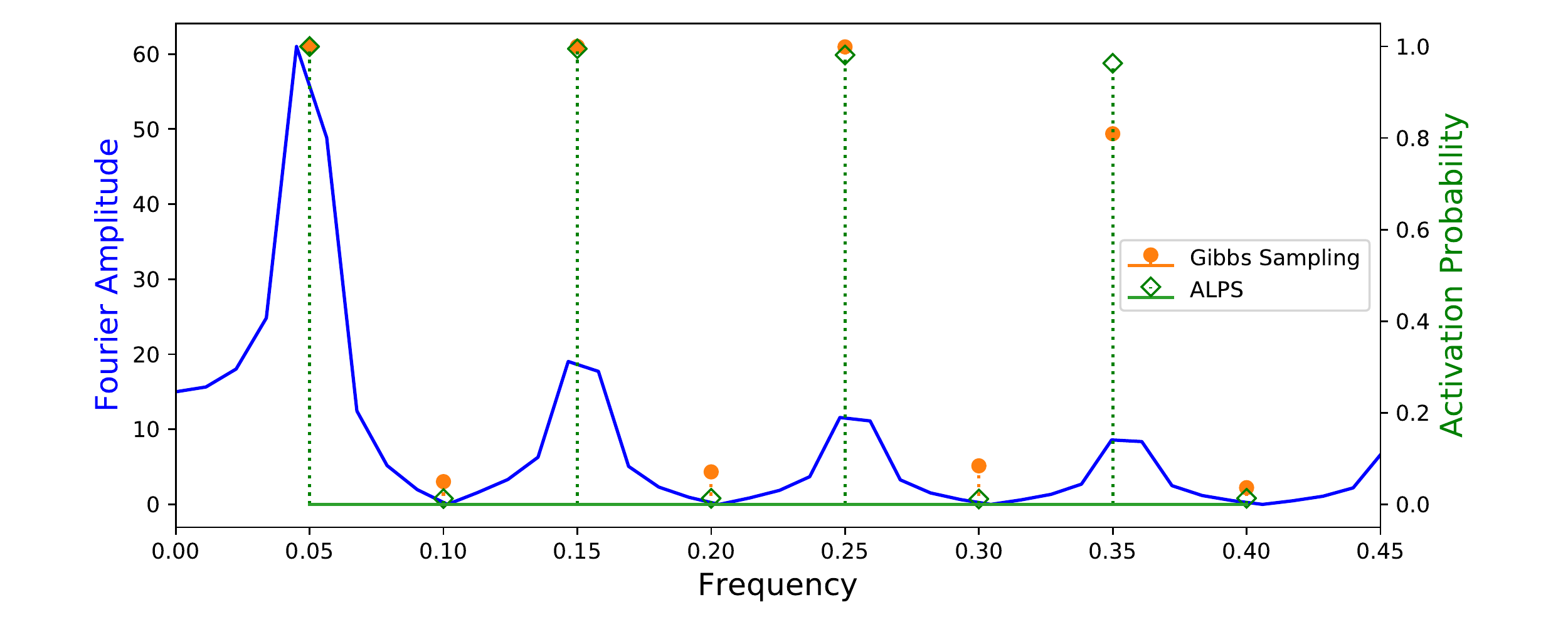}
   }
   \caption{
       Comparison between GP-ALPS and MCMC for the experiment in Section \ref{subsec:square}. 
   }
\end{figure}

We first test the model's ability to select relevant latent processes using a simple example from signal processing. We generate a single ($p = 1$) square wave of frequency $f_{\text{sq}} = 0.05$ Hz, and aim to model it as a linear combination of $m = 8$ latent GPs with linear--periodic kernels with fixed frequencies $f_i = i f_{\text{sq}}$ for $i \in {1, ..., m}$. As can be seen in Figure \ref{fig2}, GP-ALPS assigns high Bernoulli activation probabilities to the latents whose frequencies are odd multiples of $f_{\text{sq}}$, which correspond to the peaks in the square wave's power spectrum. The signal is thus reconstructed quite accurately (Figure \ref{fig1}) using the first 4 terms of the Fourier series. Furthermore, both the activation probabilities and signal reconstruction found by GP-ALPS are quite close to those obtained by sampling the exact posterior via Gibbs sampling, which indicates that the variational posterior approximates the exact one closely, despite the simplifying assumptions that have been made to enable variational inference. Interestingly, the above is true with as few as $\ell = 10$ learnable inducing points.

\subsection{Noisy Mixture of Periodic Signals}
\label{subsec:noise}

\begin{figure}[h]
  \subfigure[]{
      \label{sinusoids}
      \includegraphics[
        width=0.25\textwidth,
        height=8cm,
        clip,
        trim={0 3cm 0 3cm},
      ]{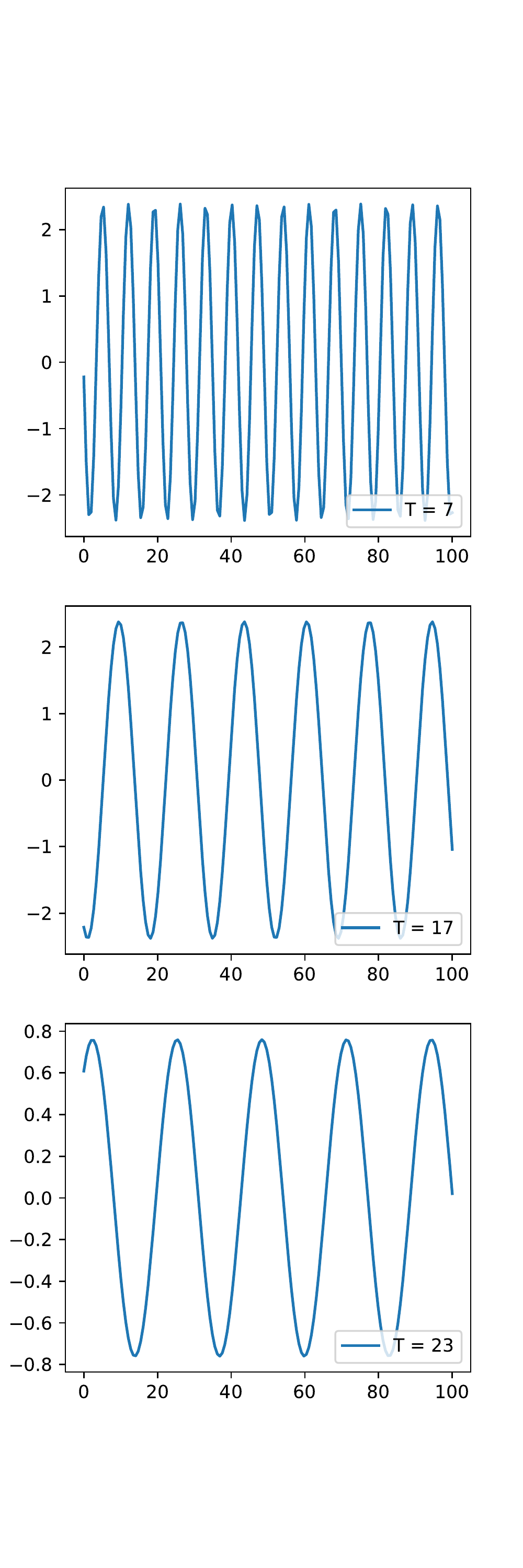}
  }
  \hfill
  \subfigure[]{
        \label{noisyoutputs}
        \includegraphics[
            width=0.7\textwidth,
            height=8cm,
        ]{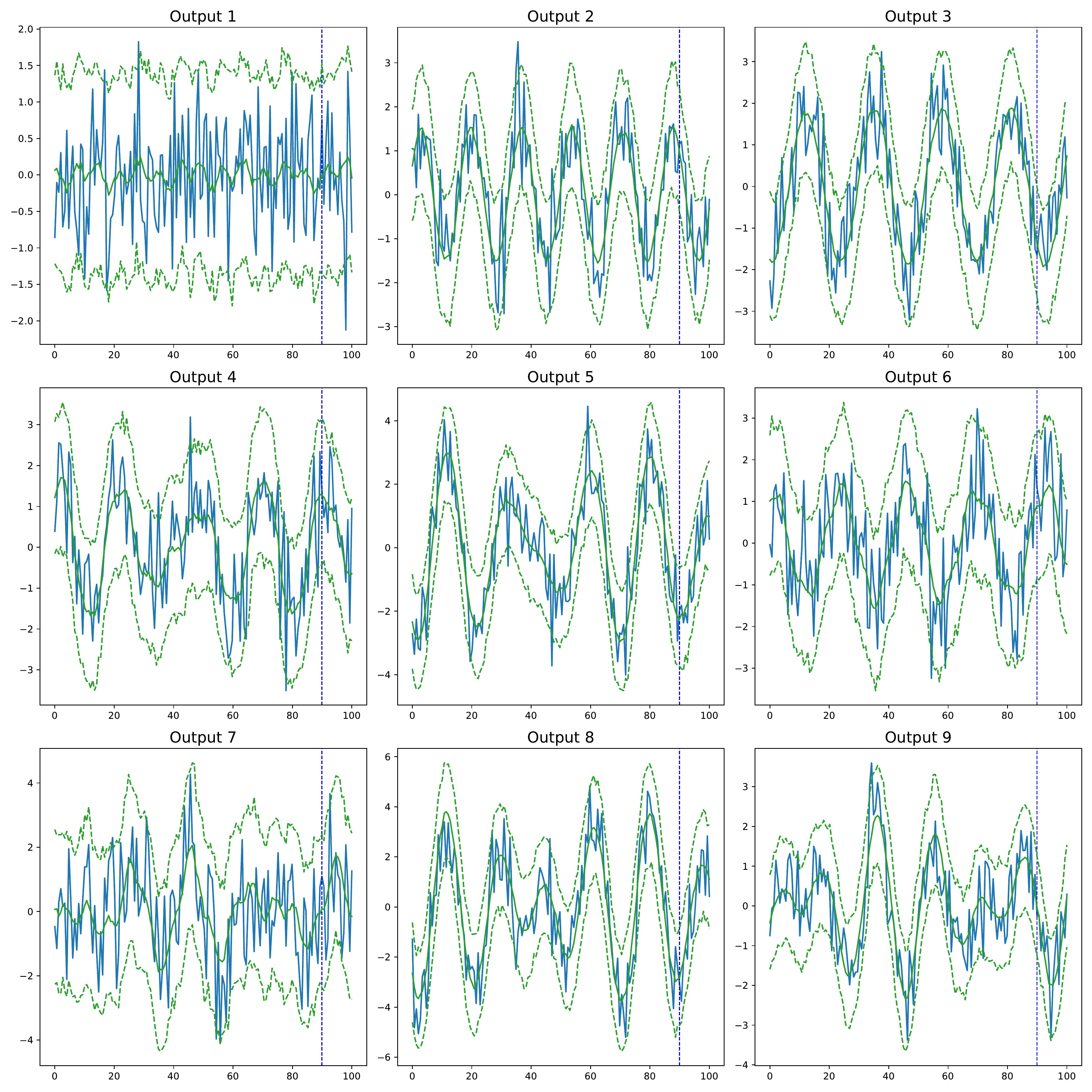}
  }
  \caption{
      Data generated for the experiment in Section \ref{subsec:noise}. \textit{(a)} Original, noiseless latent signals; \textit{(b)} outputs (blue) and predictions (green). Vertical line separates training and testing sets. 
  }
\end{figure}

In this experiment, we test the technique's ability to perform model selection in the presence of noise, as well as choose between equally good solutions. To generate the data, we start with $m^* = 3$ signals with periods $7$, $17$ and $23$ (Figure \ref{sinusoids}), then corrupt them with additive Gaussian noise and combine linearly with a fixed matrix $H^* = \big[I_3~Z\big]^T$ (where $Z \in \reals^{6 \times 6}$), to obtain $p = 9$ outputs (blue in Figure \ref{noisyoutputs}). We model the data with GP-ALPS with $m = p = 9$ linear-periodic latents, with periodicities 3, 7, 7, 11, 13, 17, 19, 23 and 23 (note the duplicates), and $\ell = 100$ learnable inducing points.

\begin{figure}[t]
  \subfigure[Mean]{
      \label{noisyHm}
      \includegraphics[width=0.28\textwidth,height=3.4cm]{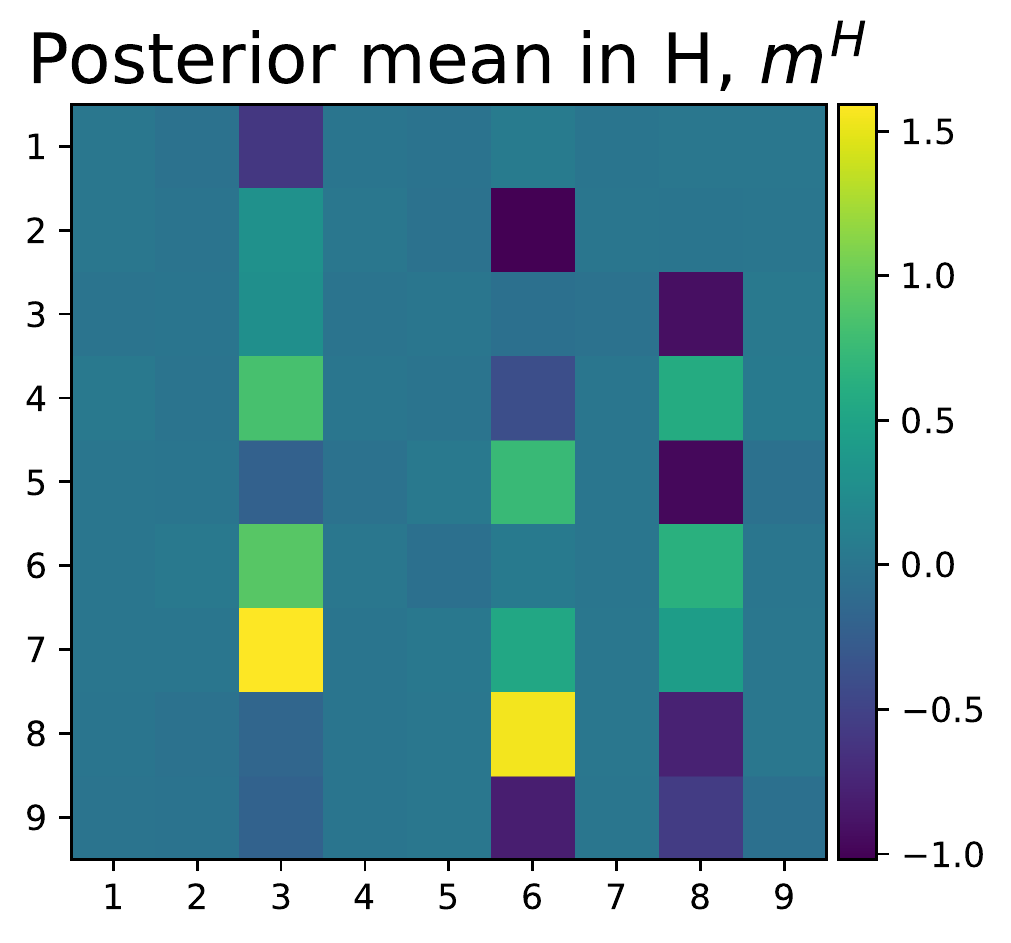}
  }
  \hfill
  \subfigure[Variance]{
      \label{noisyHv}
      \includegraphics[width=0.28\textwidth,height=3.4cm]{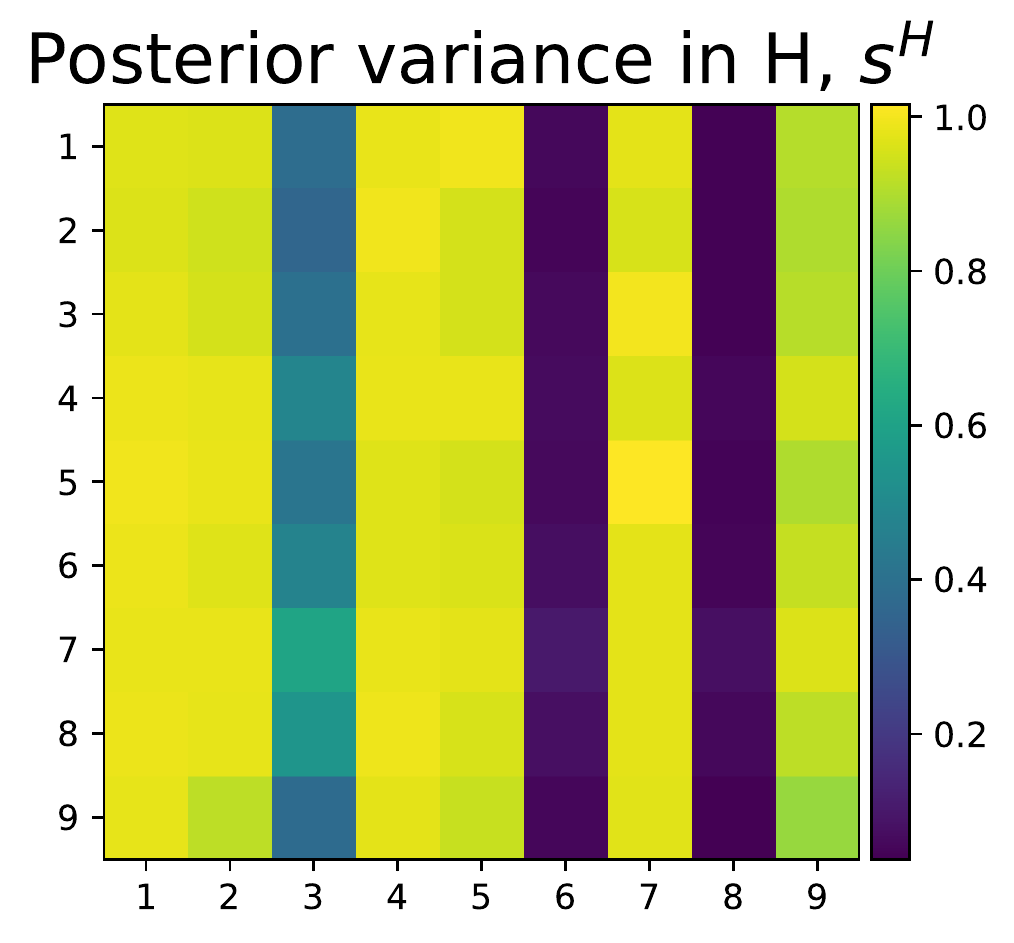}
  }
  \hfill
  \subfigure[Activation probabilities]{
        \includegraphics[width=0.36\textwidth,height=3.4cm]{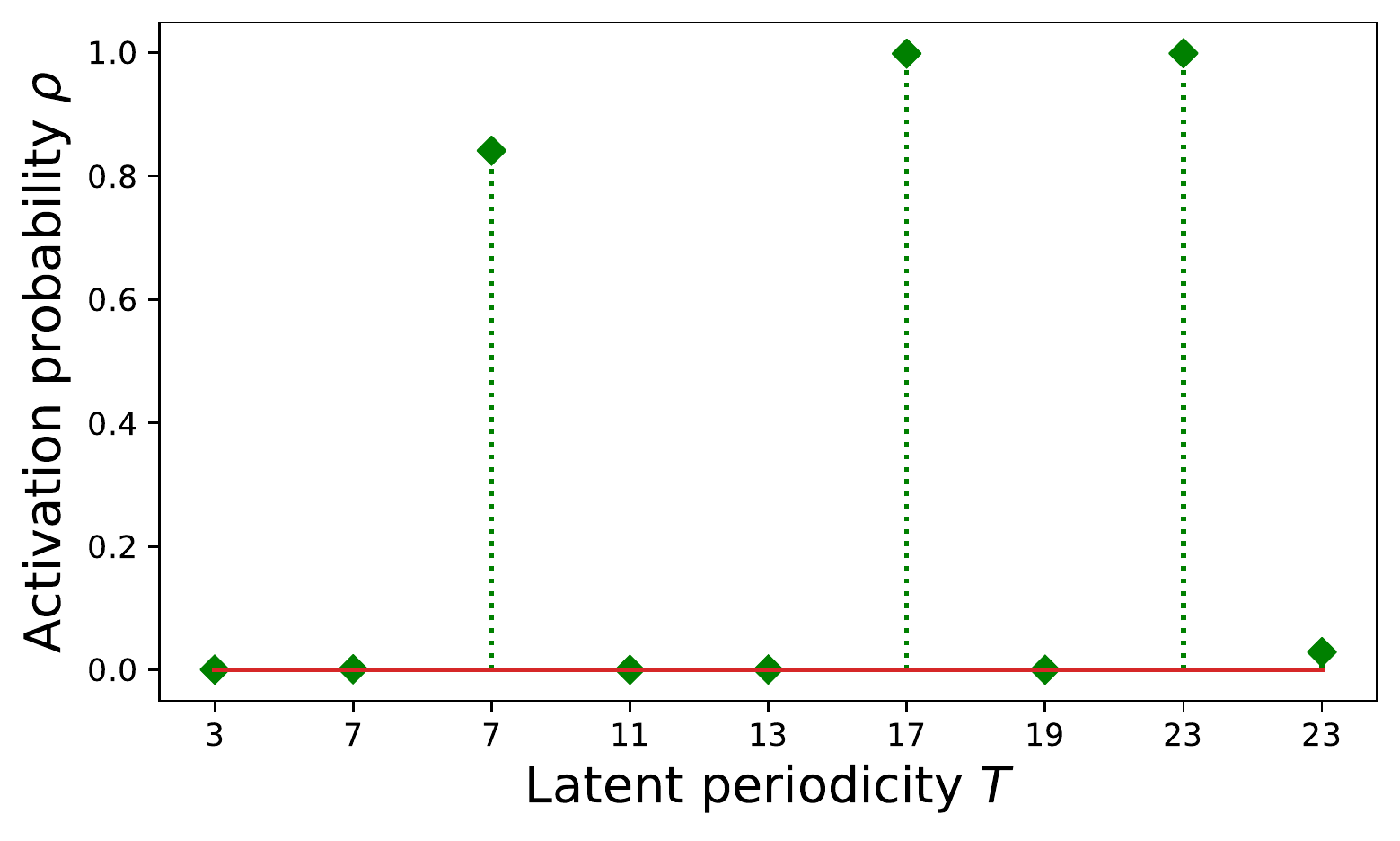}
        \label{noisyacts}
  }
  \caption{
      Approximate posteriors from the experiment in Section \ref{subsec:noise}.
  }
\end{figure}

The predictive densities are shown in green in Figure \ref{noisyoutputs}, and the trained variational posteriors $q(H)$ and $q(b)$ are shown in Figures \ref{noisyHm}, \ref{noisyHv}, and \ref{noisyacts}. GP-ALPS successfully identifies the frequencies that generated the data. Despite the noise making it virtually impossible to identify the periodicity $T = 7$ visually in the data (Output 1 in Figure \ref{noisyoutputs}), the model manages to identify its presence with high degree of certainty. Furthermore, the solution found by GP-ALPS is parsimonious---only one latent is activated for each $T = 7$ and $T = 23$.
While, intuitively, one may expect both latents with $T = 7$ (or with $T = 23$) to split the activation probability of those frequencies, this is a ``more complex'' explanation (either can be on or off) than activating only one (only one can be on or off).
By Bayesian Occam's Razor, we expect that posterior inference tends towards the simpler explanation.
Inspecting the element-wise posterior means and variances in $H$ (Figure \ref{sinusoids}), we note that elements in activated columns are estimated with low-variance Gaussians, as expected, whereas the inactive columns just revert to the standard normal prior.

\subsection{Variable Selection in Boston Housing Dataset}
\begin{figure}[t]
    \subfigure[]{
       \label{bostonact}
       \includegraphics[width=0.5\textwidth,height=4.2cm]{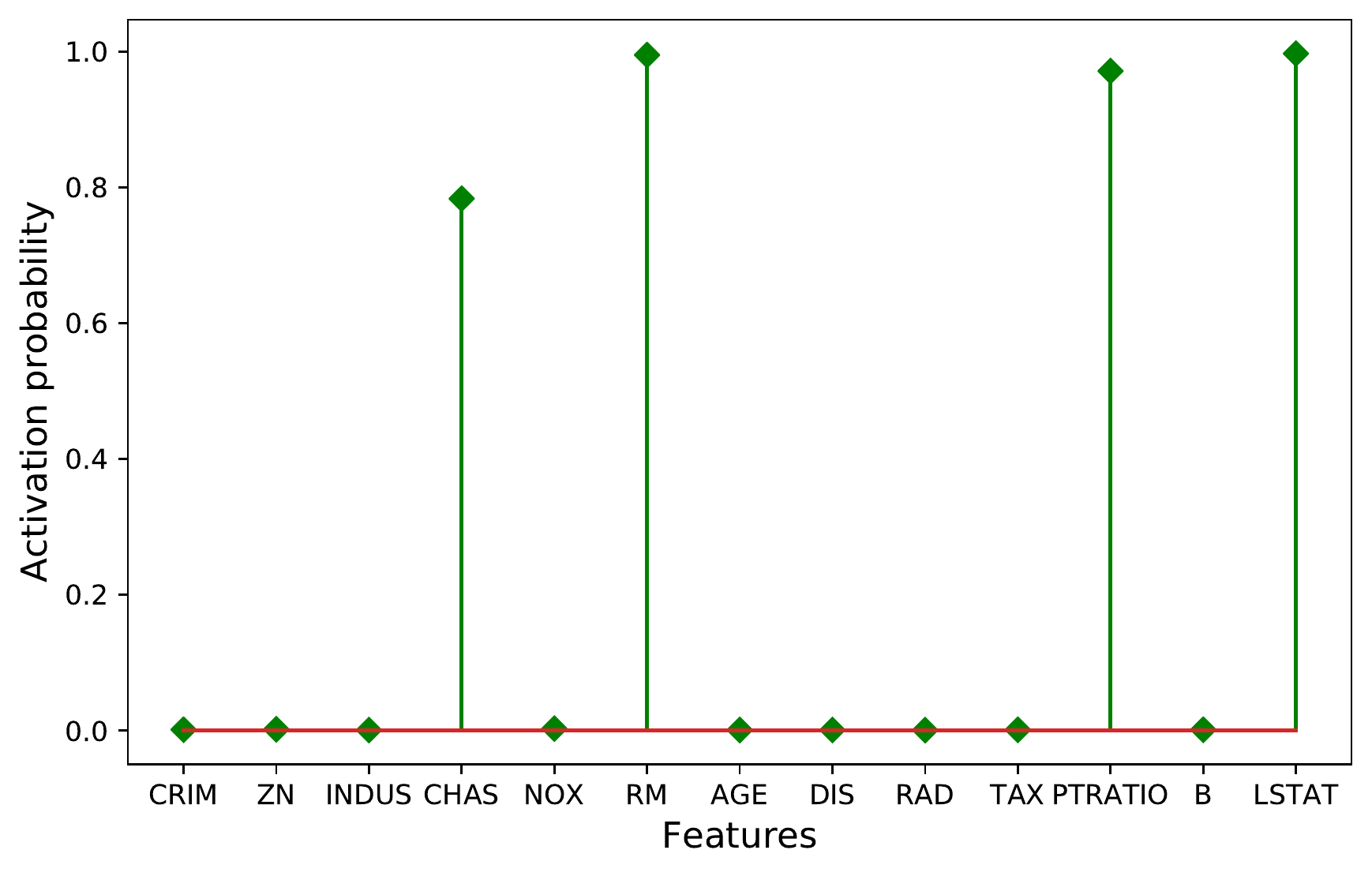}
   }
   \hfill
   \subfigure[]{
       \label{bostonhist}
       \includegraphics[width=0.5\textwidth,height=4.2cm]{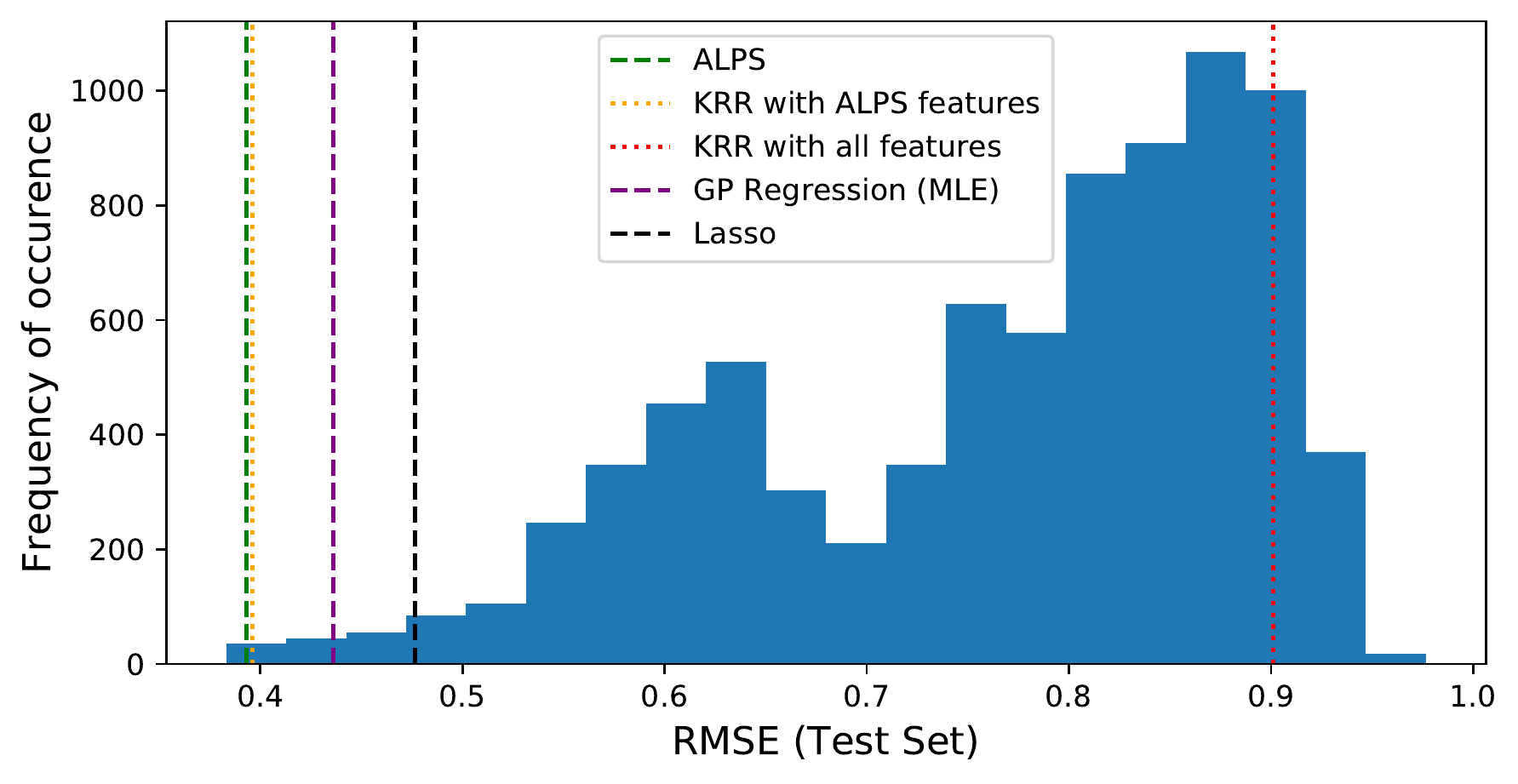}
   }
   \caption{
       (a) Feature activation probabilities found by GP-ALPS. (b) Comparison between GP-ALPS and all possible $2^{13}$ kernelised ridge regression (KRR) models on test-set RMSE.
   }
\end{figure}

Further to the experiments with synthetic data described above, we have employed GP-ALPS to perform variable selection for kernelised ridge regression (KRR), using the Boston housing dataset\footnote{
    \url{https://scikit-learn.org/stable/modules/generated/sklearn.datasets.load_boston.html\#sklearn.datasets.load_boston}
} as a motivating example. Posterior activation probabilities are shown in Figure \ref{bostonact}. Comparing the test-set results with all $2^{13} = 8192$ possible linear regression models, we demonstrate that our method performs competitively, ranking within the $0.05\%$ best models, as shown in Figure \ref{bostonhist}. This performance is comparable to the one achieved by KRR using only the features selected by GP-ALPS and superior to the one obtained by carrying regular GP regression with all features, as well as that obtained using Lasso regression. More details can be found in Appendix \ref{apd:exp}.

\clearpage

\bibliography{main}

\appendix

\clearpage

\section{Feature Selection in Boston Housing Dataset}
\label{apd:exp}
We use GP-ALPS to perform feature selection for kernelised linear regression. Our illustrative example is the Boston housing dataset, as provided in \texttt{sklearn} \citep{scikit-learn}, which contains information about properties in $506$ neighbourhoods in Boston, including median value, average number of rooms, average age and some others. The regression task is to predict the median value of a property based on $13$ other neighbourhood features.


We model the data using GP-ALPS, whereby each of the $m = 13$ latent processes corresponds to one of the input variables and the latent kernels are radial basis functions (RBF) with unit lengthscale, which is equivalent to kernelised linear regression. The number of inducing points used is $\ell = 100$, and their locations are learnt.
GP-ALPS selects $4$ out of $13$ latent processes (activation probabilities shown in Figure \ref{bostonact}) that correspond to variables CHAS (proximity to Charles river), RM (average number of rooms), PTRATIO (average pupil-teacher ratio in local schools) and LSTAT (proportion of population of lower socioeconomic status). Since the size of this data set is comparatively small, it is possible to compare the predictive performance of the model selected by GP-ALPS with that of all the other $2^{13} = 8192$ possible models. Figure \ref{bostonhist} shows the resulting histogram of test-set root-mean-squared-errors (RMSE) produced by the $8192$ kernelised linear regression models. The variable set found by GP-ALPS corresponds to the top-performing $0.05\%$ of the model space. To provide a basis for comparison, we also perform kernelised ridge regression with all 13 variables, GP regression with the kernel comprising a weighted sum of unit-lengthscale RBFs, as well as Lasso regression.

\clearpage

\clearpage
\section{Variational Inference Scheme}
\label{apd:vi}

As explained in Section \ref{sec:alps}, the generative model in GP-ALPS explains the observed signal $y(t) \in \reals^p$ as a linear-Gaussian transformation of $m$ latent Gaussian processes $x(t) \in \reals^m$, multiplied by a vector of Bernoulli variables $b$. Mathematically, this can be written down as follows:
\begin{equation*}
\begin{gathered}
    x_j \sim \mathcal{GP}(0, k_j(t, t')), \quad
    b_j \sim \operatorname{Bern}(\theta_j), \quad
    H_{ij} \sim \mathbb{}{N}(0, s_{ij}), \\
    f(t) = H (x(t) \circ b), \quad
    y_i(t) \sim \mathcal{N}(f_i(t), \sigma_i^2),
\end{gathered}
\end{equation*}
where $\circ$ refers to Hadamard product. Our goal is to compute the posterior over the latent variables, $p(X, H, b | Y)$, but this density is unfortunately intractable, so we resort to variational inference, aiming to find some distribution $q(X, H, b)$ that closely approximates the exact posterior $p(X, H, b | Y)$.

\subsection{Analytical formulation}
We start by augmenting the latent processes with \textit{inducing variables} $X^z \in \reals^{m \times \ell}$ at inducing locations $t^z \in \reals^\ell$ \citep{titsias2009variational,Hensman:2013:Gaussian_Processes_for_Big_Data,Nguyen:2014:Collaborative_Multi-Output}. This construction provides both a meaningful way of summarising the data as part of the posterior on $x$ and an efficient way of scaling to large datasets. With this addition, the evidence lower bound (ELBO) becomes:
\begin{equation*}
    \Elbo = \Expect_q \Big[ \log{\frac{\Prob(Y, X, X^z, H, b)}{q(X, X^z, H, b)}} \Big],
\end{equation*}
which we optimise numerically as in \citet{Kingma:2013:Auto-Encoding_VB}. To make this optimisation tractable, we make three important assumptions. We make a structured mean-field assumption, $q(X, X^z, H, b) = q(X, X^z) q(H) q(b)$. We take $q(X, X^z) = \Prob(X \mid X^z) q(X^z)$, as in \citet{Titsias:2009:Variational_Learning}.
For $q(X^z)$ and $q(H)$, we choose fully factorised Gaussians:
\begin{equation*}
    q(X^z) = \prod_{j = 1}^m \Normal(X^z_{j \cdot}; m^z_j, S^z_j), \quad
    q(H) = \prod_{i = 1}^p \prod_{j = 1}^m \Normal(h_{ij}; m^H_{ij}, s^H_{ij}),
\end{equation*}
where $m^z \in \reals^{m \times \ell}$, $S^z \in \reals^{m \times \ell \times \ell}$, $m^H \in \reals^{p \times m}$, $S^H \in \reals^{p \times m}$.
Since the Bernoulli distribution does not have a differentiable reparametrisation, we use a continuous relaxation of the Bernoulli distribution for $q(b)$:
\begin{equation*}
    q(b) = \prod_{j = 1}^m \text{Concrete}(b_j; \rho_j),
\end{equation*}
where $\text{Concrete}$ is the \emph{concrete} distribution \citep{maddison2016concrete}.
Here $\rho \in \small[0, 1\small]^m$.
The ELBO can then be re-written as:
\begin{equation*}
    \Elbo = \underbrace{\Expect_{q} \big(\!\log{p(Y | X, H, b)} \big)}_{\defeq \Elbo_{\text{ell}}} 
            \underbrace{- \KLD{q(X^z)}{p(X^z)}}_{\defeq \Elbo_{\text{kl}}^z}
            \underbrace{- \KLD{q(H)}{p(H)}}_{\defeq \Elbo_{\text{kl}}^H}
            \underbrace{- \KLD{q(b)}{p(b)}}_{\defeq \Elbo_{\text{kl}}^b}
\end{equation*}
Let us consider each of the terms above in turn.

\subsection{Expected log-likelihood ($\Elbo_{\text{ell}}$)}
Start from the full expression for the expected log-likelihood:
\begin{align*}
    \Elbo_{\text{ell}} & = \Expect_{p(X | X^z) q(X^z) q(H) q(b)} \big[ \log{p(Y | X, H, b)} \big].
\end{align*}
Since the conditional likelihood inside the expectation does not depend on $X^z$, we first marginalise it out, similarly to \citet{dezfouli2015scalable}:
\begin{equation*}
    q(X) = \int dX^z p(X | X^z) q(X^z) =  \prod_{j = 1}^m \underbrace{\Normal(X_j; A_j m^z_{j}, \tilde{K}_j + A_j S_j A_j^T)}_{q(X_j)}
\end{equation*}
where $A_j = K_{cz}^j (K_{zz}^j)^{-1}$ and $\tilde{K}^j = K_{cc}^j - A_j K^j_{zc}$, such that $K_{cc}^j$, $K_{cz}^j$ and $K_{zc}^j$ are Gram matrices constucted using latent kernel $k_j(\cdot, \cdot)$ on input vectors $t$ and $t^z$. Adapting Theorem~1 from \citet{dezfouli2015scalable} to our parametrisation of $q(X^z)$, we then write the ELL as:
\begin{align*}
    \Elbo_{\text{ell}} = \Expect_{q(H)q(b)} \Big[ \sum_{t = 1}^n \Expect_{q(x_t)} \big[ \log{p(y_t | x_t, H, b)} \big] \Big],
\end{align*}
where $q(x_t) = \Normal(x_t; d_t, S^x_t)$ such that $(d_t)_j = (A_j m^z_j)_t$ and $(S^x_{t})_{jj} = \tilde{K}^j_{tt} + (A_j)_t^T S^z_j (A_j)_t$. The final expression for ELL we use is then:
\begin{align*}
    \Elbo_{\text{ell}} = \underbrace{\sum_{t = 1}^n \Expect_{q(H)q(b)q(x_t)} \Big[ \log{p(y_t | x_t, H, b)} \Big]}_{\defeq \Elbo_{\text{ell}}^{(t)}},
\end{align*}
whose gradients we compute using the reparametrisation trick \citep{Kingma:2013:Auto-Encoding_VB, rezende2014stochastic, titsias2015local} on variational posteriors $q(x_t)$, $q(H)$ and $q(b)$.

\subsection{KL-divergence in activation variables ($\Elbo_{\text{kl}}^b$)}
To compute
\begin{equation*}
    \Elbo^{b}_{\text{kl}} = - \Expect_{q(b)} \Big[ \log{\frac{q(b)}{p(b)}} \Big],
\end{equation*}
we also approximate $p(b)$ with the concrete distribution.
Again, we use the reparametrisation trick to compute gradients.

\subsection{KL-divergence in inducing variables ($\Elbo_{\text{kl}}^z$)}
Both $q(X^z)$ and $p(X^z)$ are block-diagonal multivariate Gaussians, so the KL-term has a closed analytical form:
\begin{align*}
    \Elbo_{\text{kl}}^z & = -\KLD{q(X^z)}{p(X^z)} = - \sum_{j = 1}^m \KLD{q(X^z_j)}{p(X^z_j)} \\
                        & = - \sum_{j = 1}^m \KLD{\Normal(X^z_j; m_j^z, S_j^z)}{\Normal(X^z_j; 0, K^j)} \\
                        & = - \frac{1}{2} \sum_{j = 1}^m \Big[ \text{Tr}\big[(K^j)^{-1} S^z_j \big] + (m^z_j)^T (K^j)^{-1} m^z_j - \ell + \log{\frac{|K^j|}{|S_j^z|}} \Big],
\end{align*}
so gradients can be computed analytically or by automatic differentiation.

\subsection{KL-divergence in mixing matrix ($\Elbo_{\text{kl}}^H$)}
Both $q(H)$ and $p(H)$ are also diagonal multivariate Gaussians, so the KL-term is simply
\begin{align*}
\Elbo_{\text{kl}}^H & = -\KLD{q(H)}{p(H)} = - \sum_{i = 1}^p \sum_{j = 1}^m \KLD{q(H_{ij})}{p(H_{ij})} \\
                    & = - \sum_{i = 1}^p \sum_{j = 1}^m \KLD{\Normal(H_{ij}; m^H_{ij}, s^H_{ij})}{\Normal(H_{ij}; 0, s_{ij})} \\
                    & = - \sum_{i = 1}^p \sum_{j = 1}^m \Big[ \frac{1}{2} \log{\frac{s_{ij}}{s_{ij}^H}} + \frac{s_{ij}^H + (m_{ij}^H)^2}{2s_{ij}} - \frac{1}{2} \Big],
\end{align*}
which, again, can be differentiated analytically or using automatic differentiation.

\subsection{Summary of the optimisation problem}
All in all, the variational objective we aim to maximise is:
\begin{align*}
    \Elbo = \Elbo_{\text{ell}} + \Elbo_{\text{kl}}^b + \Elbo_{\text{kl}}^z + \Elbo_{\text{kl}}^H.
\end{align*}
Observing that $\Elbo_{\text{ell}}$ is a sum over data points, and that other terms do not depend on observations $Y$, the ELBO will also be a sum over data points:
\begin{align*}
    \Elbo = \sum_{t = 1}^n \Elbo^{(t)} = \sum_{t = 1}^n \Big[ \Elbo_{\text{ell}}^{(t)} + \frac{1}{n} \big( \Elbo_{\text{kl}}^b + \Elbo_{\text{kl}}^z + \Elbo_{\text{kl}}^H \big) \Big],
\end{align*}
thus, the objective is amenable to stochastic gradient-based optimisation, which is helpful for scaling the model to large datasets.

\subsection{Temperature of the concrete distributions}
\begin{figure}[h]
    \centering
    \includegraphics[
        width=0.5\textwidth,
    ]{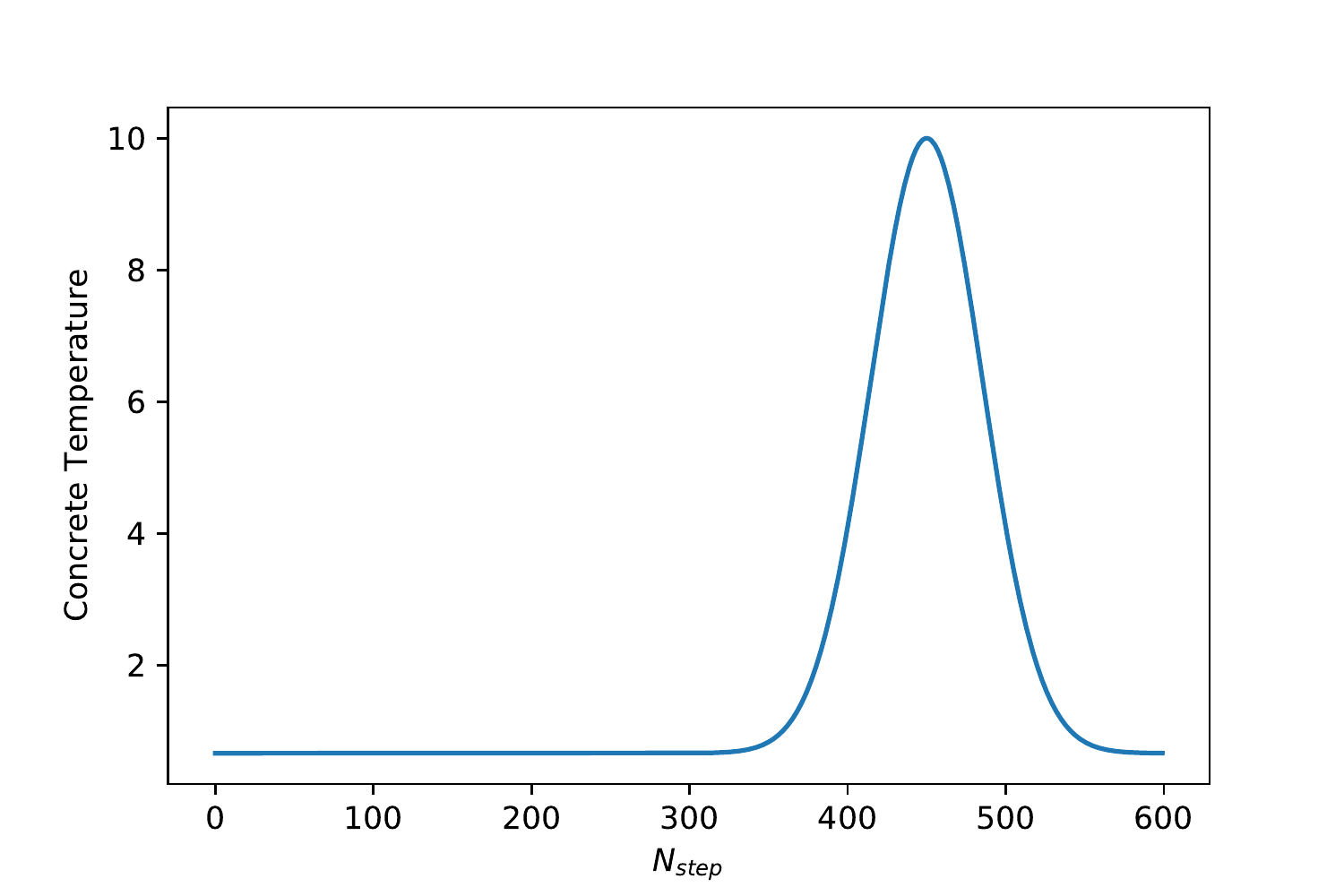}
    \caption{
        Visualisation of the annealing scheme for the temperature of the concrete distributions
    }
    \label{fig:temp}
\end{figure}

For the temperature of the concrete distributions in $q(b)$, we use the following annealing scheme:
\[
     T(n, N)
     =
     0.66 + (10.0 - 0.66) \exp\left(
        -\frac
            {(n - 0.75 N)^2}
            {0.083^2 N^2}
     \right)
\]
where $n$ is the current iteration and $N$ is the total number of iterations.
This annealing scheme is visualised in Figure \ref{fig:temp}.
The idea behind the temperature starting low is that the rest of the parameters can be optimised before latent processes start being dropped out. As for the temperature parameter in the continuous relaxation of $p(b)$, it is chosen to be $1 / 2$, as in \citet{maddison2016concrete}.

\clearpage

\clearpage

\section{Markov Chain Monte Carlo}
\label{apd:mcmc}
This appendix summarises the derivations of the conditionals needed to perform Gibbs sampling from the intractable exact posterior of GP-ALPS. As stated in Section \ref{subsec:VI}, the following three conditionals are of interest:
\[
    p(X | Y, H, b), \quad
    p(H | X, Y, b), \quad \text{and} \quad
    p(b_j | X, Y, H, b_{\neg j}).
\]
Let us consider each of them in turn.

\subsection{$p(X | Y, H, b)$}
Start by writing down the Bayes' theorem:
\begin{align*}
p(X | Y, H, b) & \propto p(Y | X, H, b) ~ p(X) = \Normal(y; H_b x, S_b)~\Normal(x; 0, K),
\end{align*}
where $y = \text{vec}(Y)$, $x = \text{vec}(X)$, $H_b = \big(H~\text{diag}(b)\big) \otimes I_n$, $S_b = \text{diag}({\sigma^2}) \otimes I_n$ and $K$ is the block-diagonal multi-output kernel matrix. Note that the above is a Bayesian linear regression problem with a Gaussian prior, so the posterior form is well-known \citep[p.~93]{Bishop:2006:Pattern_Recognition_and_Machine_Learning}:
\begin{align*}
p(x | y, H_b) = \Normal(x; S_x H_b^T S_b^{-1}y, S_x),
\end{align*}
where $S_x = \big[K^{-1} + H_b^T S_b^{-1} H_b\big]^{-1}$.

\subsection{$p(H | X, Y, b)$}
As before, writing down the Bayes' theorem:
\begin{align*}
    p(H | Y, X, b) & \propto p(Y | H, X, b) ~ p(H) = \Big[\prod_{i = 1}^p p(y_{i} | X, h_i, b) \Big] \Big[ \prod_{i = 1}^p p(h_i) \Big] \\
                       & = \prod_{i = 1}^p \Big[p(h_i) p(y_{i} | X, h_i, b) \Big] = \prod_{i = 1}^p \Big[~\Normal(h_i; 0, s_{ij})~\Normal(y_i; X^T~ \text{diag}(b)~h_i, \sigma^2_i I_n)~\Big],
\end{align*}
where $h_i \in \reals^m$ is $i^{\text{th}}$ row of H, and $y_i \in \reals^n$ is the $i^{\text{th}}$ output. The above amounts to  $p$ independent Bayesian linear regression problems, so, as before, the posterior form is well-known \citep[p.~93]{Bishop:2006:Pattern_Recognition_and_Machine_Learning}:
\begin{align*}
p(h_i | y_i, b, X) = \Normal(h_i; S_h~\text{diag}(b)~Xy_{i} / \sigma_i^2, S_h),
\end{align*}
where $S_h = \big[\frac{1}{s} I_m + \frac{1}{\sigma^2_i}~\text{diag}(b)~XX^T \text{diag}(b) \big]^{-1}$.

\subsection{$p(b_j | X, Y, H, b_{\neg j})$}
Writing down the Bayes theorem for each $b_j$:
\begin{align*}
p(b_j | Y, X, H, b_{\neg j}) & \propto p(Y | X, H, b)~p(b_j)  = p(b_j) \Big[ \prod_{i = 1}^p \prod_{t = 1}^n p(y_{ti} | b, x_t, h_i) \Big] \\
                                  & \propto \exp{\Big[ b_j \log\frac{\theta_j}{1 - \theta_j} \Big]} \prod_{i = 1}^p \prod_{t = 1}^n \exp{\Big[-\frac{(y_{ti} - \sum_{k = 1}^m h_{ik} b_k x_{tk})^2}{2 \sigma^2_i}\Big]} \\
                                  & = \exp{\Big[ b_j \log\frac{\theta_j}{1 - \theta_j} \Big]} \prod_{i = 1}^p \prod_{t = 1}^n \exp{\Big[ - \frac{1}{2 \sigma_i^2} \big( y_{ti}^2 - 2 y_{ti} \sum_{k = 1}^m h_{ik} b_k x_{tk} + (\sum_{k = 1}^m h_{ik} b_k x_{tk})^2 \big) \Big]} \\
                                  & = \exp{\Big[ b_j \log\frac{\theta_j}{1 - \theta_j} \Big]} \prod_{i = 1}^p \prod_{t = 1}^n \text{exp} \Big[ - \frac{1}{2 \sigma_i^2} \big( y_{ti}^2 - 2 y_{ti} \sum_{k \neq j} h_{ik} b_k x_{tk} - 2 y_{ti} h_{ij} b_j x_{tj} \\
                                  & + (\sum_{k \neq j} h_{ik} b_k x_{tk})^2 + 2 h_{ij} b_j x_{tj} + (h_{ij} b_j x_{tj})^2 \Big] \\
                                  & \propto \exp{\Big[ b_j \log\frac{\theta_j}{1 - \theta_j} \Big]} \text{exp} \Big[ -\frac{b_j}{2} \sum_{i = 1}^p \frac{1}{\sigma^2_i} \big(2 h_{ij} \sum_{t = 1}^n \sum_{k \neq j} h_{ik} b_k x_{tj} x_{tk} \\
                                  & + h_{ij}^2 \sum_{t = 1}^n x^2_{tj} - 2 h_{ij} \sum_{t = 1}^n x_{tj} y_{ti} \big)\Big] \\
                                  & = \exp{\Big[ b_j \big( \log\frac{\theta_j}{1 - \theta_j} + c_j \big) \Big]} = \Bern \big[b_j; \frac{\theta_j e^{c_j}}{e^{c_j} + \theta_j} \big],
\end{align*}
where
\begin{align*}
    c_j = -\frac{1}{2} \sum_{i = 1}^p \frac{1}{\sigma^2_i} \big(2 h_{ij} \sum_{t = 1}^n \sum_{k \neq j} h_{ik} b_k x_{tj} x_{tk} + h_{ij}^2 \sum_{t = 1}^n x^2_{tj} - 2 h_{ij} \sum_{t = 1}^n x_{tj} y_{ti} \big).
\end{align*}

\end{document}